\documentclass[10pt,journal,compsoc]{IEEEtran}

\usepackage{amsmath,amssymb,amsfonts}
\usepackage{graphicx}
\usepackage{ragged2e}
\usepackage{textcomp}

\usepackage[ruled,vlined]{algorithm2e}
\usepackage{caption, subcaption}

\usepackage{todonotes}
\usepackage[para]{footmisc}
\usepackage{booktabs}
\usepackage{xcolor,colortbl}
\usepackage{hyperref}
\usepackage{multirow}
\usepackage{cleveref}
\usepackage{rotating}
\usepackage{wrapfig}
\usepackage[noadjust]{cite}
\usepackage{paralist}
\usepackage{listings}
\usepackage{lipsum}

\begin{document}

\title{Question Answering Over Biological Knowledge Graph via Amazon Alexa}

\author{
\IEEEauthorblockN{
	Md. Rezaul Karim\IEEEauthorrefmark{2}\IEEEauthorrefmark{1},
	Hussain Ali\IEEEauthorrefmark{2}, 
	Prinon Das\IEEEauthorrefmark{2}, 
	Mohamed Abdelwaheb\IEEEauthorrefmark{2},
	Stefan Decker\IEEEauthorrefmark{2}\IEEEauthorrefmark{1} 
}\\ \vspace{2mm}

    \IEEEauthorblockA{\IEEEauthorrefmark{2} \scriptsize{Computer Science 5 - Information Systems and Databases, RWTH Aachen University, Germany}}\\
        \IEEEauthorblockA{\IEEEauthorrefmark{1} \scriptsize{Fraunhofer Institute for Applied Information Technology FIT, Germany}}
}

\IEEEtitleabstractindextext{%
\begin{abstract}
\justifying
    Structured and unstructured data and facts about drugs, genes, protein, viruses, and their mechanism are spread across a huge number of scientific articles. These articles are a large-scale knowledge source and can have a huge impact in disseminating knowledge about mechanisms of certain biological processes. A knowledge graph~(KG) can be constructed by integrating such facts and data and be used for data integration, exploration, and federated queries. However, exploration and querying large-scale KGs is tedious for certain group of users due to lack of knowledge about underlying data assets or semantic technologies. A question answering~(QA) system allows answer natural language questions over KGs automatically using triples contained in a KG.   
    Recently, the use and adaption of digital assistants is getting wider owing to their capability at enabling users voice commands to control smart systems or devices. 
    This paper is about using Amazon Alexa's voice-enabled interface for QA over KGs. As a proof-of-concept, we use the well-known \emph{DisgeNET} KG, which contain knowledge covering 1.13 million gene-disease associations between 21,671 genes and 30,170 diseases, disorders, and clinical or abnormal human phenotypes. Our study shows how Alex could be of help to find facts about certain biological entities from large-scale knowledge bases. 
\end{abstract}

\begin{IEEEkeywords} Question answering, Knowledge graphs, Ontology, Semantic web, Bioinformatics, Digital assistants, Amazon Alexa. \end{IEEEkeywords}}
\maketitle

\IEEEdisplaynontitleabstractindextext
\IEEEpeerreviewmaketitle

\IEEEraisesectionheading{\section{Introduction}\label{sec:introduction}}

\IEEEPARstart{D}{omain} experts are often interested in gathering and comprehending knowledge and mechanism of certain biological process, e.g., diseases to design strategies in order to develop prevention and therapeutics decision making process. \textit{``Knowledge is something that is known and can be written down''}~\cite{nonakatakeuchi1995}. Knowledge containing simple statements, e.g., \textit{``TP53 is an oncogene"} or quantified statements, such as \textit{``All oncogenes are responsible for cancer"} can be extracted from structured sources such as knowledge or rule bases. Moreover, knowledge can be extracted from external sources like scientific articles, where KG could be an effective means to capture facts from heterogeneous data sources. For example, scientific literature and patents provide a huge treasure of structured and unstructured information about different biological entities. One prominent example is PubMed, which contain millions of scientific articles is a great source of knowledge in biomedical domain~\cite{xu2020building}. PubMed data are mostly unstructured and heterogeneous. This makes the knowledge extraction process very challenging. 

The problem of semantic heterogeneity is further compounded due to the flexibility of semi-structured data and various tagging methods applied to documents or unstructured data. Owing to SW technologies that offer functionality to connect previously isolated pieces of data and knowledge, associate meaning to them, and represent knowledge extracted from them. In particular, ontology-based named entity extraction and disambiguation help with unambiguous identification of entities in heterogeneous data and assertion of applicable named relationships that connect these entities together. Semantic Web~(SW) addresses data variety, by proposing graphs as a unifying data model, to which a data can be mapped in the form of a graph structure. A graph may not only contain data, but also metadata and domain knowledge~(ontologies containing axioms or rules), all in the same uniform structure, and are then called knowledge graph~(KGs)~\cite{wilcke2017knowledge,hogan2020knowledge}. 

A simple statement can be accumulated as an edge in a KG, while quantified statements provide a more expressive way to represent knowledge, which however requires \textit{ontologies}~\cite{hogan2020knowledge}. Hogan et al.~\cite{hogan2020knowledge} defined KG as \textit{a graph of data intended to accumulate and convey knowledge of the real world, whose nodes represent entities of interest and whose edges represent potentially different relations between these entities}. Nodes in a KG represent entities and edges represent binary relations between those entities~\cite{hogan2020knowledge}. A KG can be defined as $G=\{E,R,T\}$, where $G$ is a labelled and directed multi-graph, and $E, R, T$ are the sets of entities, relations, and triples, respectively and a triple can be represented as $(u,e,v) \in T$, where $u \in E$ is the head node, $v \in E$ is the tail node, and $e \in R$ is the edge connecting $u$ and $v$~\cite{hogan2020knowledge}. 

However, building a domain KG has several core requirements, such as formal conceptualization to indicate the logical design of the KG depicted by a specific, predefined domain-specific ontology, and the modelling of domain knowledge, represented by semantically interrelated entities and relations~\cite{abu2020domain}. Ontologies are semantic data models that define the types of things that exist in a domain and the properties that can be used to describe them, including the relationships between them~\cite{hogan2020knowledge}. An ontology not only defines the relationships between concepts~\cite{hitzler2009foundations}, but also provides a formal representation of domain-specific entities\cite{alirezaie2019semantic}. 

Information extraction~(IE) is the process of automatically extracting structured knowledge and facts from such unstructured and/or semi-structured documents or electronically represented sources~\cite{Liddy.2001}. IE is typically divided into named entity recognition~(NER), entity linking, and relation extraction. Relation extraction also involves relation classification, which is typically formulated as a classification problem to classify the relationship between the entities identified in the text~\cite{xue2019fine}. A classifier takes a piece of text and two entities as inputs and predicts possible relations between the entities as output. Once instances are extracted, they can be stored as Resource Description Framework~(RDF)\footnote{In RDF, the linking structure of a graph forms a directed graph and triples are represented in the form of $(subject,predicate,object)$.} triples, where each triple forms a connected component of a sentence for the KG. A number of languages have been proposed for querying RDF data~\cite{hogan2020knowledge}, including the SPARQL query language\footnote{SPARQL is the protocol/language to query RDF, which allows querying not only over graph data but also between disparate graphs. Link: \url{https://www.w3.org/TR/sparql11-query/}}, the Cypher Query Language\footnote{\url{https://neo4j.com/developer/cypher/}}, Gremlin Query Language\footnote{\url{https://docs.janusgraph.org/basics/gremlin/}}. 

Reasoning over KGs enables consistency checking to recognize conflicting facts, classification by defining taxonomies, and deductive inferencing by revealing implicit knowledge from a set of facts~\cite{futia2020integration}. Further, \textit{deductive reasoning} can be used to entail extended knowledge such as \textit{``TP53 is responsible for cancer"}\cite{karim_phd_thesis_2022}. However, a large-scale KG can have billions of linked entities expressing their relationships, where each node represent an entity and each edge signifies a semantic relationship between entities~\cite{karim2019drug}. These makes the exploration, processing, and analysis of large-scale KGs pose a great challenge to current computational methods. To provide cancer diagnosis reasoning over the DNN models, an integrated domain-specific KG is required, which is subject to the availability of an efficient NLP-based information extraction method and a domain-specific ontology~\cite{karim_phd_thesis_2022}. 

In this paper, we report a case study of using digital assistants -- in particular Amazon Alexa's voice-enabled interface for QA over KGs. We use the well-known \emph{DisgeNET} KG, which contain knowledge covering 1.13 million gene-disease associations between 21,671 genes and 30,170 diseases, disorders, and clinical or abnormal human phenotypes. Our study shows how Alex could be of help to find facts about certain biological entities from large-scale knowledge bases. 
The rest of the paper is structured as follows: \Cref{sec:rw} critically reviews related works. \Cref{sec:method} describes the proposed approach in details, covering construction of black-box models, training, interpreting the black-box model, and generation of decision rules and local explanations. \Cref{sec:results} illustrates experiment results, including a comparative analysis with baseline models on all datasets. \Cref{sec:con} summarizes this research with potential limitations and points some possible outlook. 

\section{Related Works}\label{sec:rw}
\noindent Research initiatives are gradually adopting SW technologies~\cite{karim2018improving} such as KGs, knowledge bases~(KBs), and domain-specific ontologies as the means of building structured networks of interconnected knowledge~\cite{futia2020integration}. Hogan et al.~\cite{hogan2020knowledge} have provided a comprehensive review of articles on KGs, covering knowledge graph creation, enrichment, quality assessment, and refinement. Apart from these literature that focuses on the theoretical concepts, several large-scale KGs have been constructed either by manual annotation, crowd-sourcing~(e.g., DBpedia) or by automatic extraction from unstructured data~(e.g., YAGO\footnote{\url{https://github.com/yago-naga/yago3}})~\cite{wang2015explicit} targeting KG analytics for specific use cases. 
Life sciences is an early adaptor of SW technologies. Scientific communities have focused on constructing large-scale KGs for life science research. For example, Bio2RDF\footnote{\url{https://github.com/MaastrichtU-IDS/bio2rdf} and PubMed KG~\cite{xu2020building}} are developed to accelerate bioinformatics research. The former integrates 35 life sciences datasets such as dbSNP, GenAge, GenDR, LSR, OrphaNet, PubMed, SIDER, WormBase, contributing 11 billion RDF triples. 

Alshahrani et al.~\cite{alshahrani2017neuro}, built a biological KG based on the gene ontology~(GO), human phenotype ontology~(HPO), and disease ontology. Then they performed feature learning over the KG. Their method combines knowledge representation using symbolic logic and automated reasoning, with neural networks to generate embeddings of nodes. The learned embeddings are used in downstream application development such as link prediction, finding candidate genes of diseases, protein-protein interactions, and drug target relation prediction. Hasan et al.~\cite{hasan2020knowledge} developed a prototype KG based on the Louisiana Tumor Registry dataset\footnote{https://sph.lsuhsc.edu/louisiana-tumor-registry/}. Their approach provides scenario-specific querying, schema evolution for iterative analysis, and data visualization. Although the resultant KG found effective at population-level treatment sequences, it does not provide comprehensive knowledge about cancer genomics for the majority of the cancer types as it is built on a limited data.


Although numerous work focused on data integration, building KGs, and querying over biological domain, only a few works have been focused on question answering and retrieving information from KGs via voice-enabled interface. Other works focus on improving the quality and enriching multiple KGs in order to query via voice-enabled devices~\cite{10.1145/3366423.3380289}. Haase et al.~\cite{Haase2017AlexaAW} focused on using wikidata to answer factual questions via Alexa. The self-attention and cross-attention mechanism and factoring in information about KGs to perform entity alignment, which entails determining which elements of different graphs refer to the same ``entities'' shows how it will be helpful for voice-enabled interfaces~\cite{amazonscience}. The idea is to improve computational efficiency while at the same time improving performance, speeding up graph-related tasks such as question answering via Alexa.

\section{Methods} \label{sec:method}
To make the information retrieval from large-scale KGs having higher layers and intricacies and to develop a proof-of-concept for Amazon Alexa's voice-enabled interface, we use the well-known \emph{DisgeNET} KG, which contain knowledge covering 1.13 million gene-disease associations between 21,671 genes and 30,170 diseases, disorders, and clinical or abnormal human phenotypes. 
The database covers gene-disease associations~(GDAs), disease-disease associations~(DDAs), and variant-disease associations~(VDAs). 
Disease gene identification is a process by which experts identify the mutant genotype responsible for an inherited genetic disorder. This dataset provides robust coverage towards our task but does not mean it has a RDF triples for all the diseases and genes out there. Association triples exist only if disease and gene/variant is supported by evidence from external sources. 

Extracting data is always tied to a SPARQL query, which admittedly is not the most user friendly interface. Therefore, the straight forward solution is to implement some APIs which will be accompanied by a front-end interface,for example voice -enabled Alexa device and tied with the back-end, and then, KGs will serve as a fully functional application on its own, which is certainly a viable solution. The DisgeNET dataset has lot of potentials, allowing developing hypothesis and answering very interesting questions. However, we focused on following range of questions only:

\begin{itemize}
  \item What genes are associated with a specific disease?
  \item What genes are associated with several diseases?
  \item What diseases are associated with a protein class?
  \item How many publications~(e.g., PubMed or Cancer Index) are there that support gene-disease associations?
  \item Are there evidence of specific gene-disease association?
\end{itemize}

\subsection{Knowledge graph integration and accessibility}
\noindent For easy data exploration and querying, communication between the KG and Alexa is established in several steps. 
First an SPARQL endpoint is configured for public data access. 
Finally, we automate the CI/CD pipeline in which the artefacts re configured such that it outputs an image with custom dataset at build time following a run test doing a CURL request for checking the status of the Jena server. 

\begin{figure}
    \centering
    \includegraphics[width=0.5\textwidth]{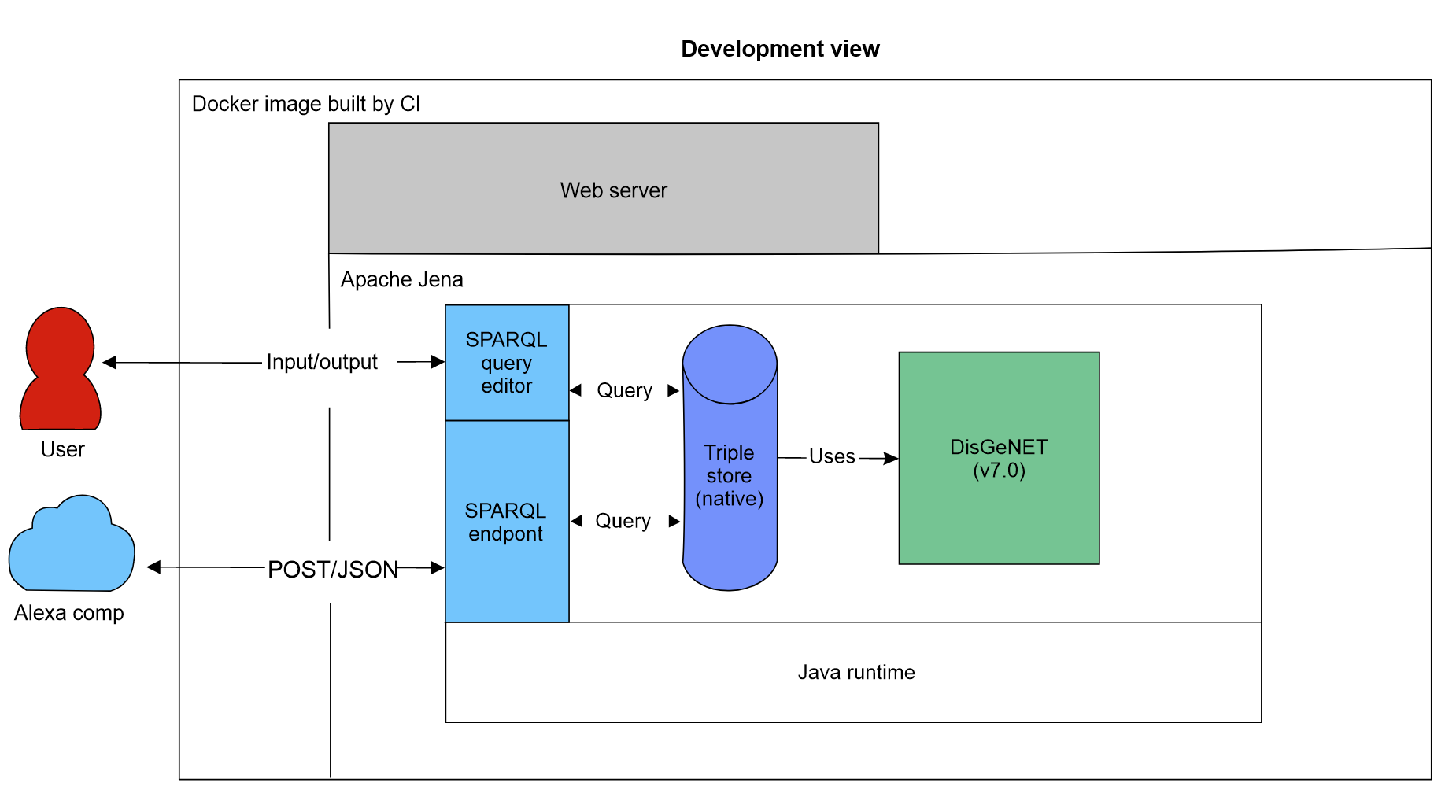}
    \caption{Alexa and knowledge graph integration}
    \label{fig:arch}
\end{figure}

\subsection{Introduction to Alexa for for developers} 
\noindent We aim at interacting via natural language, thereby allowing users query via text or voice. Alexa is Amazon’s cloud-based voice service available on devices both from Amazon and third-party device manufacturers. Alexa offers tools, APIs, and documentations to make it easy to build custom tools for Alexa, called ``Skills''. Therefore, we aim at building a set of custom skills that would act as the interface with the KG. 
We experimented with- and used the following components of Alexa for developing Alexa skills related to the KG: 

\begin{enumerate}
    \item \textbf{Alexa management console} - for managing the lambda function~(LF). The skill in the Alexa development console is linked to the LF through a specific skill id, where the computation of entire skill is executed end-to-end. 
    \item \textbf{Alexa developer console} - is used for writing the main codes along with the development of the interaction model and testing and building the skill components.
    \item \textbf{Cloudwatch logs} - is used for monitoring and logging.
    \item  \textbf{Alexa ask CLI console}  - used for developing the skills in a command line interface. However, unlike in Alexa developer console where there is a scope of creating multiple files, all the python scripts were encapsulated in a single script for smooth execution of all the skills\footnote{This problem is attributed to the ASK sdk framework being still in its early stage and has not been properly integrated to address all bugs.}.
\end{enumerate}

\begin{figure*}
    \centering
    \includegraphics[width=\textwidth]{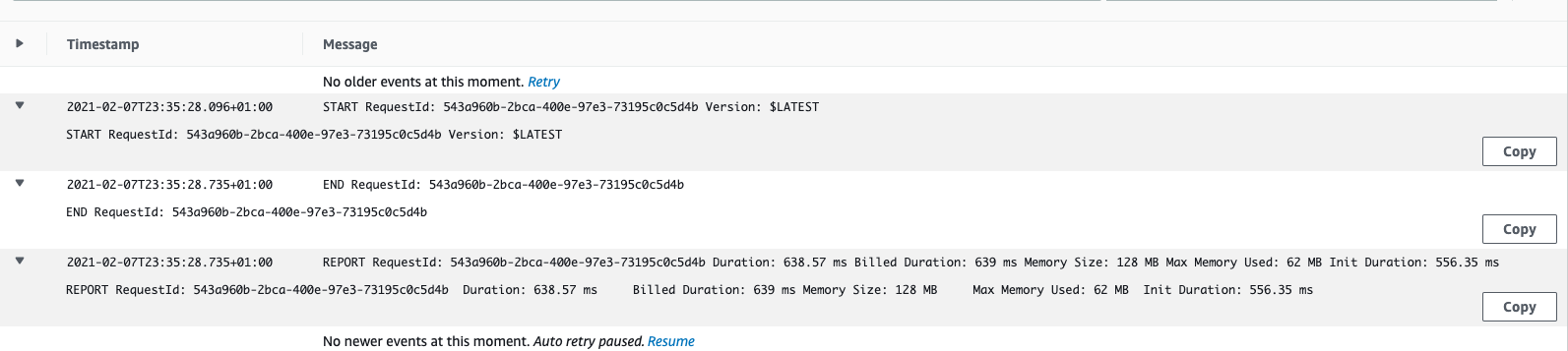}
    \caption{Monitoring and logging through Cloudwatch logs}
    \label{fig:build_skill}
\end{figure*}


The development of entire skill on developer console has two major module components:

\begin{enumerate}
    \item The \textbf{interaction model} in the Amazon developer website which has the automated speech recognition and natural language understanding components.
    \item The \textbf{intent handlers} which are run by the Amazon web services\footnote{Via the lambda function.} uses the Alexa skill kit trigger.
\end{enumerate}

\begin{figure*}
    \centering
    \includegraphics[width=\textwidth]{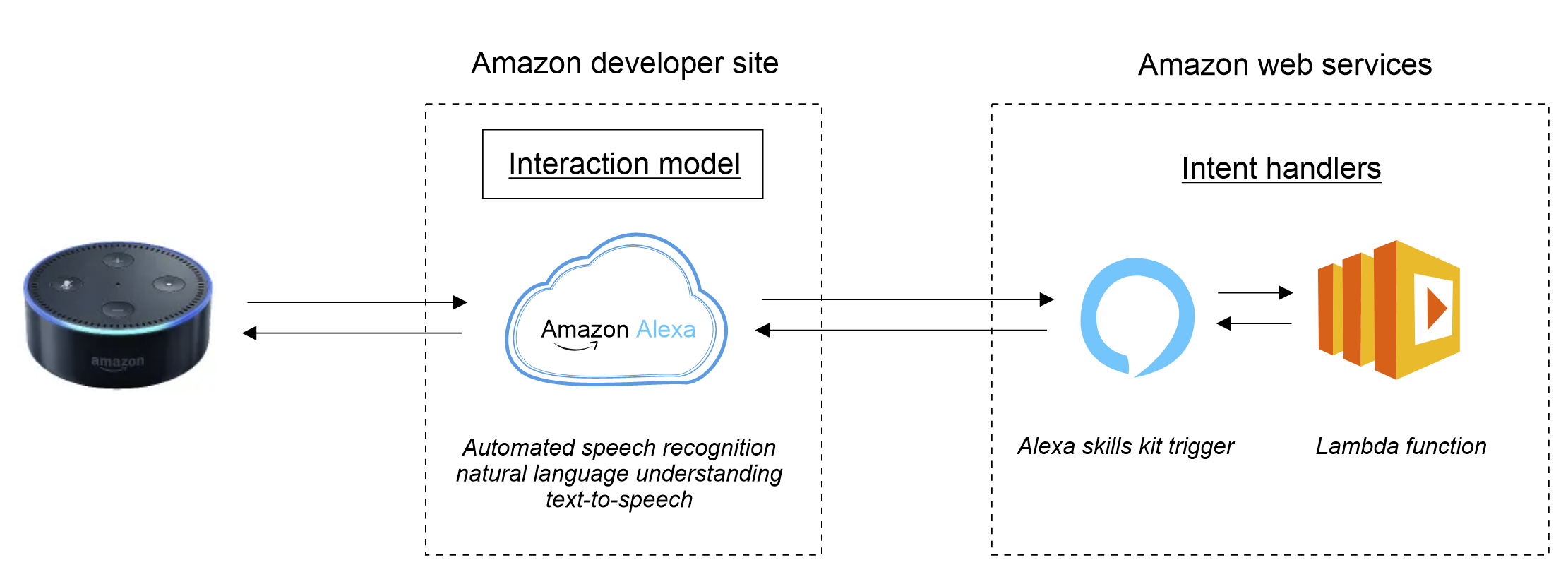}
    \caption{Alexa developer component modules high level view, conceptually recreated based on literature~\cite{highlevelview}}
    \label{fig:build_skill2}
\end{figure*}

\subsection{Alexa skill components} 
 \noindent We define the terms of the components of the skill, which help us understand the architecture, and design choices that we took when implementing our skills. First, a set of \emph{`Intents'} is created that represents as set of actions that the user can do with the skill. Besides, they are used to represent the code functionality of the skill. A set of samples \emph{`Utterances'} is created that specify the phrases user can say to call or invoke those above-mentioned intents. In order to map each utterance to the appropriate intent, this mapping creates the \emph{`interaction model'} of the skill. This interaction model serves as the backbone of the skill. An \emph{`invocation name'} is created that uniquely identifies each skill that the user must include to start a conversation with the skill.
 
Although, a set of images, audio, and video files that can be included in the  skill, they must be stored publicly on an accessible site, i.e., each item of those must have a unique URL. However, for the purposes of this skill, we did not use any media files. A \emph{`cloud-based service'} that takes those intents as requests and then takes the suitable action, it acts as an API controller for our skill. Moreover, it must be accessible over the internet, and we provide the endpoint to our skill during the initial configuration. Finally a configuration that brings all those things together, and for that we can always use Alexa developer console, or we can do internally and use \emph{`ask-cli'} later for deployment.

\subsection{Building knowledge graph skill}
\noindent The KG skills are built in several steps.  
First, a user asks the voice-enabled interface of Alexa for a query. Then the request is activated through an invocation name mentioned by the user, which essentially activates the lambda function associated with the skill in the background. The request is then enters the language interaction model of the Alexa, where it is converted from speech to text. The received text is then represented into JSON serialization format, which then flows to our customized handler code where the code recognizes which intent is appropriate to handle the query. 

The appropriate handler has the RDF query, which calls the API through SPARQL and gets back the appropriate response.
he return result is formatted properly and converted to proper JSON output. The output is converted to Speech Synthesis Markup Language~(SSML), which is the synthetic audio of the result. SSML is an XML-based markup language for speech synthesis applications. Finally, the synthetic audio is played back through Alexa to the user giving the appropriate answer.

\begin{figure*}
    \centering
    \includegraphics[width=0.8\textwidth]{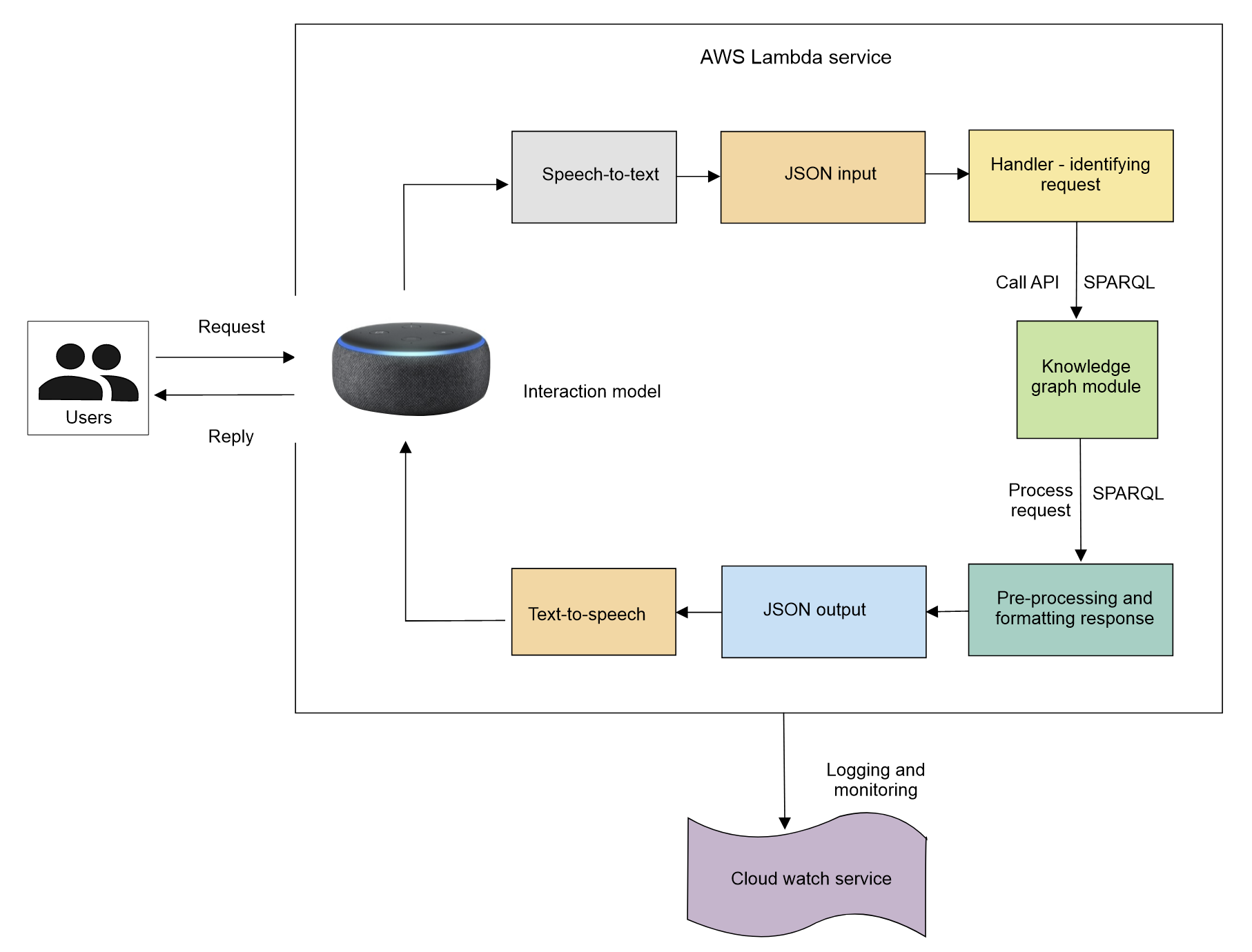}
    \caption{Over architecture of the knowledge graph skill used by Alexa}
    \label{fig:kg_skill}
\end{figure*}

\subsection{Building Alexa skills} 
\begin{figure}
    \centering
    \includegraphics[width=0.4\textwidth]{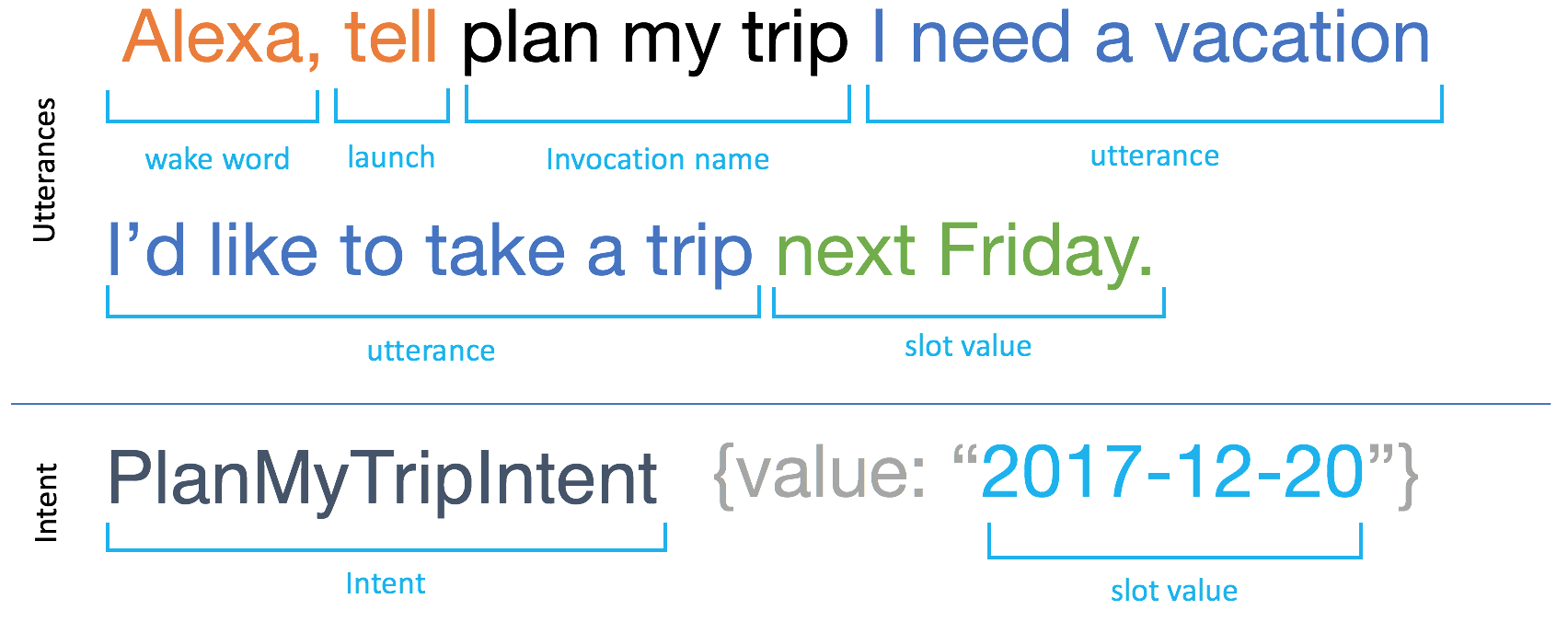}
    \caption{Utterances and intents~(based on literature~\cite{whatuserssay})}
    \label{fig:planMyTripUtteranceAndIntent}
\end{figure}

\noindent The process of building a skill, is best defined through a pipeline of different processes built upon each other incrementally as building blocks upon which we get a better picture of a fully operational and published skill. We approach building the skills such that a user will ask a question about a disease for example, Alexa will then respond with appropriate answer, giving him/her some information about the subject that he/she asked about, after that Alexa will wait for the next question then rinse and repeat. Then it is extended to dialog system, with re-prompt from Alexa to expand on certain points if the user so wishes, but first a proof of concept of the first point needed to be fulfilled. The overall process of building Alexa skills is depicted in ~\cref{fig:build_skill}, outlining the architecture of a custom Alexa skill, what are the building blocks, and how do they connect, then we can move onto our interpretation of a custom skill. \Cref{fig:planMyTripUtteranceAndIntent} shows the for a visual representation, and examples, that can aid to clarify things up.

First, we start by defining the invocation name of our skill, which as mentioned acts as the identifier for the skill. This consists of two parts, first is what is call wake word, which is something along the lines of calling Alexa to get it starting, for example ‘Hey Alexa’, or just saying ‘Alexa’. It’s like calling someone to get his/her attention, and this is shared among all Alexa skills and we don’t need to define it in the configuration, but we need to say/type it when testing or using the skill. Now, the second part is the invocation name, which is unique to each skill, in such case we name our skill, or just write as the main action of our skill. Combined, we can say for example  \textit{`Hey Alexa, open cooking skill'},  or  \textit{`Hey Alexa, plan my trip'}. In our case our invocation name is ‘Language team’, and you invoke the skill by saying  \textit{`Hey Alexa, open language team'}.

\begin{figure}
    \centering
    \includegraphics[width=0.5\textwidth]{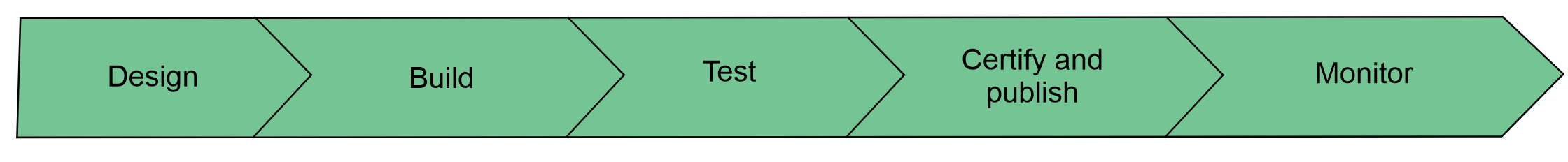}
    \caption{Skill building process recreated based on~\cite{customskill}}
    \label{fig:build_skill3}
\end{figure}

As of utterances, it describes what the user wants. For intents, it is the intention of the phrase, so a user will have a question in mind, or request, but the way he/she phrases it is different and can be said in a lot of different ways, but they all have the same purpose, we always like to use the example of capitals and countries, because its intuitive and easy to understand. Therefore, if we are to create a geography skill, which would answer questions about our lovely planet. One of its capabilities is knowing the capital of countries in the world. A user will approach the skill wanting to know the capital of France for example. 

Now, this is the intent of the user, at the end of the day, whatever one utters out of his/her mouth, or types out, he needs to know what is the capital of France. Here comes the tricky part, because in this simple case we can have multitude of different utterances, which describes the phrase itself, for example: \textit{`What is the capital of France?'},  \textit{`Show me the capital of France'}, \textit{`I want to know the capital of France'}, or even simply ` \textit{Capital of France?'}. The task of the developer is mapping all these utterances to the intent to getting the capital of a given country.

Since defining all these utterances for every single country is a tedious even for a simple task covering 250+ countries and each having 5+ different utterances, the number of utterances will be immense, and hugely inefficient to define, and store. Therefore, we did not cover all the utterances. As of slots, they act as the variable for our intent so instead of the intent being getting the capital of France, the intent becomes getting the capital of {country}, and all the utterances reflect that: ‘What is the capital of {country}, ‘Show me the capital of {country}, `I want to know the capital of {country}, or even simply `Capital {country}?'. In addition to  utterances to intent mapping, the developer must define slots, and map those to utterances and intents, this whole model becomes what is known as interaction model. Further, we will give more in depth analysis into each one along with examples, from our own skill.

As shown in \cref{fig:planMyTripUtteranceAndIntent}, we present of a full interaction with Alexa, this interaction can be done either by typing in the interface, or by just simply speaking to the Alexa device. The user starts invoking Alexa by saying the invocation name, which like mentioned contains both the wake word plus the invocation name of the skill itself, the user says, `Alexa tell plan my trip', now Alexa is ready and waiting for whatever utterance comes next, but now Alexa knows that whatever comes next is related to the ‘plan my trip’ skill. The next is the `utterance', where the user says what he/she need. Here, the user wants to plan a vacation, so he/she says, `I’d like to take a trip …'. The skill maps this utterance to the appropriate intent. However, the skill is waiting for slot value. 
If the user says, `next Friday', which is the slot value that we need to fulfil the intent. Finally with all that information combined, we have our full `PlanMyTripInent' intent with its slot value `2017-12-20'. At this point what we need to do is to pass the intent call and intent values to the intent handler, which will then either be passed to an external back-end engine, or handled internally. 

We start by creating our own Alexa skill, it is quite simple all we needed is going to the Alexa developer console with our own Amazon account. And then we defined the invocation name of our skill. We choose it as ‘Language Team’, this was a reference to the sub team of this specific project, to invoke our skill a user can simply say ‘Alexa open Language Team’. Next up, was choosing our platform for development, and here we will take this opportunity to speak about the background configuration. The handlers and logic of the code are hosted and complied on Amazon Lambda, and the Alexa skill ID is used a reference and connector between both. Besides, we use the cloud watch to monitor the incoming logs. 

\section{Evaluation}\label{sec:results}
\noindent During the development of the proof-of-concept, we carried out several experiments\footnote{\url{https://github.com/rezacsedu/Question-answering-over-Amazon-Alexa}}. We briefly cover our findings. 

\subsection{Datasets} 
\noindent The DisGeNET~(v7.0) dataset contains 1,134,942 GDAs, between 21,671 genes and 30,170 diseases, disorders, traits, and clinical or abnormal human phenotypes, and 369,554 VDAs between 194,515 variants and 14,155 diseases, traits, and phenotypes. The construction of DisgeNET is a potent resource for this task. It provides a large number to study gene and disease together which follows from \cref{table:1}, i.e over 1135045 gene-disease associations GDAs which is between 21671 genes and 30170 diseases, disorders, traits, and clinical or abnormal human phenotype. Besides, DisGeNET is integrated to Linked Open Data~(LOD) cloud providing cross-referencing among other datasets.

\begin{table*}
    \centering
    \begin{tabular}{ |p{3cm}|p{2cm}|p{2cm}|p{2cm}| p{2cm}| }
         \hline
         \multicolumn{5}{|c|}{Dataset statistics~(\# of triples)} \\
         \hline
          Source type & Genes & Diseases & Associations & Evidences\\
         \hline
         Curated   & 9703 & 11181 & 84038 & 137822\\
         Inferred   & 13258 & 14843 & 233738 & 313885\\
         Animal models & 3334 & 3171 & 16660 & 22171\\
         Literature    & 18898 & 18171 & 858354 & 2738700\\
         \hline
         ALL & 21671 & 30170 & 1134942 & 3178358\\
         \hline
    \end{tabular}
    \caption{DisgeNET RDF triple count for v7.0}
    \label{table:1}
\end{table*}

\begin{figure}
    \centering
    \includegraphics[width=0.5\textwidth]{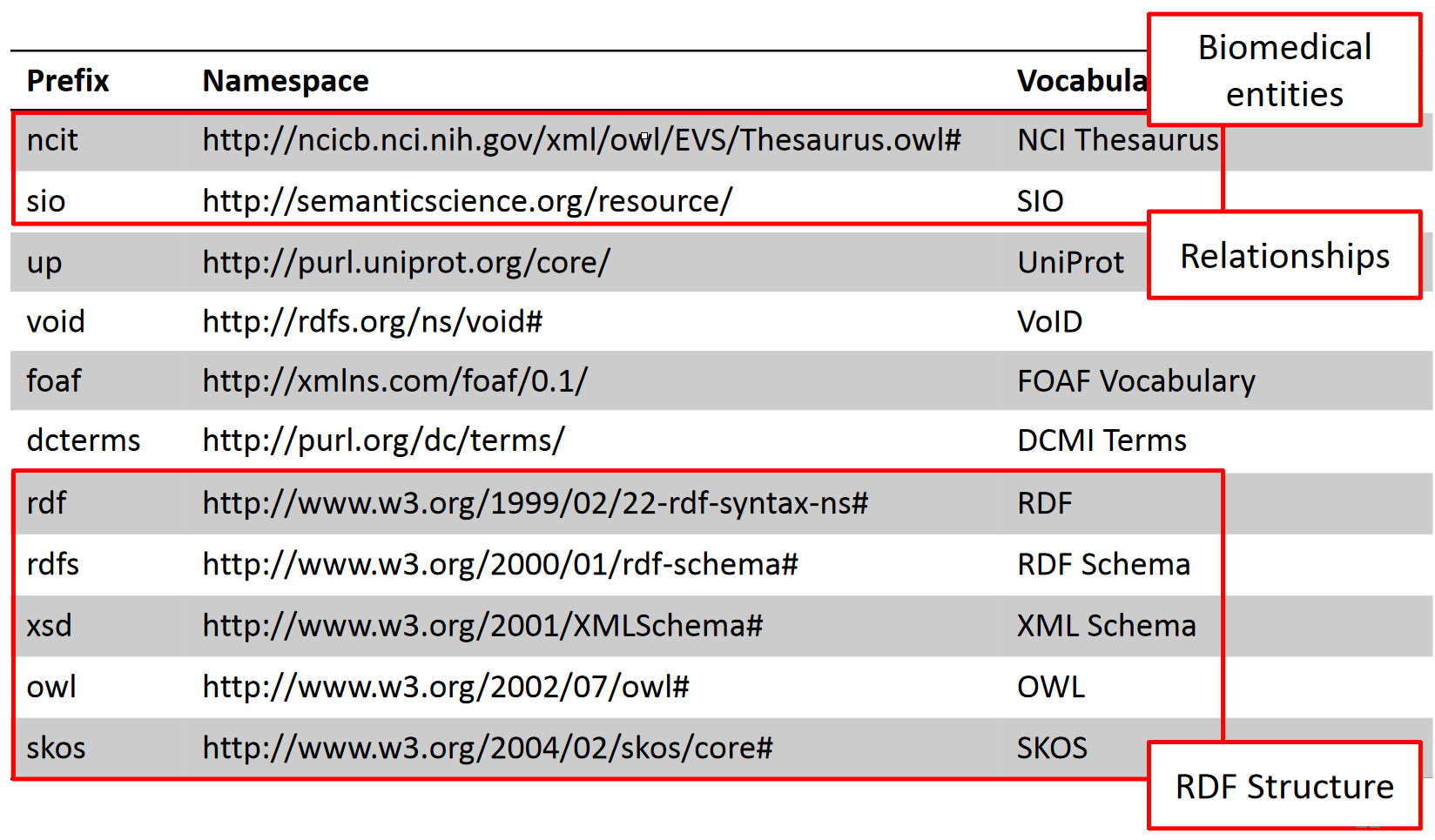}
    \caption{DisGeNET standard ontology table~(source:~\cite{tutorial})}
    \label{fig:std-ontologies}
\end{figure}

\begin{figure}
    \centering
    \includegraphics[width=0.5\textwidth]{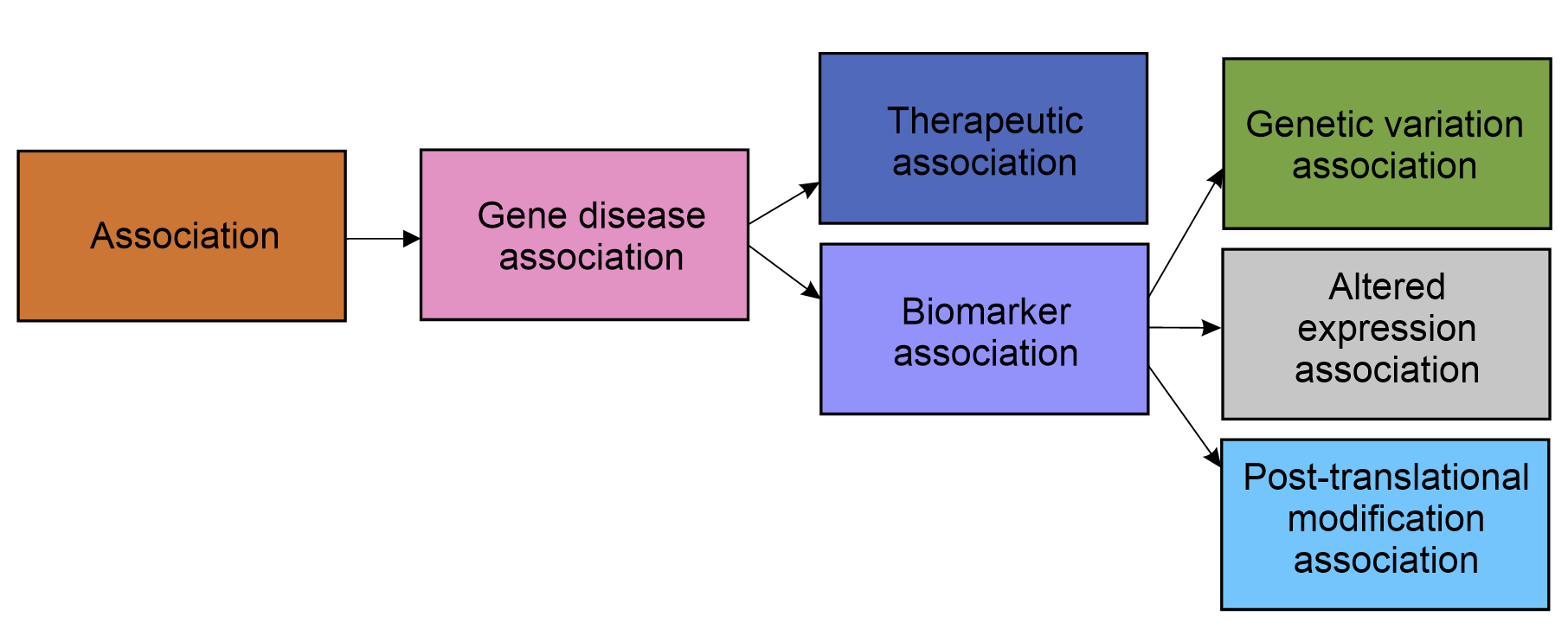}
    \caption{DisgeNET association type ontology~(source:~\cite{wikiwand})}
    \label{fig:1}
\end{figure}

For information retrieval, understanding the ontology is crucial. The association of DisGeNET ontology is shown in \cref{fig:1}. It defines or categorizes the association into two main types namely Therapeutic Association and Biomarker Association where the former indicates that the gene/protein has a therapeutic role in the amelioration of the disease and the latter indicates that the gene/protein either plays a role in the aetiology of the disease~(e.g. molecular mechanism that leads to disease) or is a biomarker for a disease.
    
This entity binds the Gene-Disease Identification as the `Gene-Disease Association', which has properties which is the PubMed article if any as supporting evidence, the source entity which tells about the vocabulary source it is taken from in the LOD identifying the dataset and its version. A score calculated by the Jaccard similarity measure and an SNP identifier. This concludes one entity, although there are other external sources that can be referenced. 



\subsubsection{What is the supporting evidence?}
\begin{figure*}
    \centering
    \includegraphics[width=0.8\textwidth]{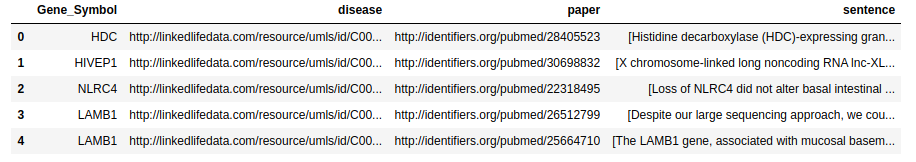}
    \caption{Excerpts supporting the associations}
    \label{fig:excerpts}
\end{figure*}

The question can arise upon the credibility for the GDA which is profound in the database with the help of excerpts that metadata around the association which provides the not only the link to the article at least one along with the sentence in string format which the Alexa system can read out immediately to the user. The results of the query is shown in \cref{fig:excerpts}.

\begin{figure}
\scriptsize{
    \begin{lstlisting}[frame=bt,numbers=none,language=SPARQL,linewidth=\linewidth,morekeywords={FILTER,AS,GROUP,BY,VALUES,PREFIX}]
        SELECT DISTINCT ?gSymbol as ?Gene_Symbol ?disease ?paper ?sentence
        WHERE {
            ?gda sio:SIO_000628 ?gene, ?disease ;
             sio:SIO_000772 ?paper ;
             dcterms:description ?sentence .
            ?gene a ncit:C16612 ;
                sio:SIO_000205 ?symbolUri .
            ?symbolUri dcterms:title ?gSymbol .
            ?disease a ncit:C7057
        }
        LIMIT 50
\end{lstlisting}}
    \caption{SPARQL query for list of oncogenes responsible for any type of cancer cited in Cancer Index} 
    \label{fig:sparql_query_1}
\end{figure}

\subsubsection{Can we find the top-k GDA?}
\begin{figure*}
    \centering
    \includegraphics[width=0.8\textwidth]{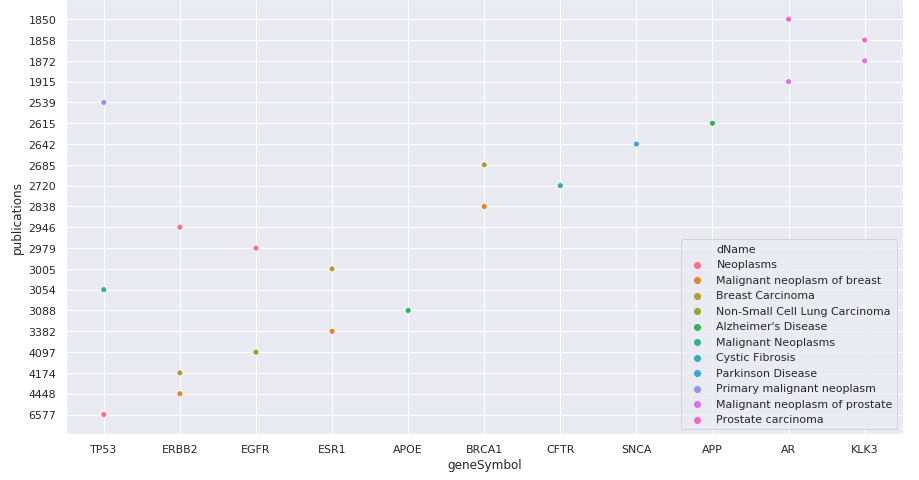}
    \caption{Top-20 gene-disease associations}
    \label{fig:top20}
\end{figure*}

\noindent For an aerial view over the association, it is often of interest for experts to know about the popular and most researched causality that is the disease in result of the gene or for some set of genes found in a person experts can tell what could be the most likely disease to occur if not already. For this type of query, there has to be a criterion which in our analysis was to set to the number of PubMed articles supporting the results. In \cref{fig:top20}, one can see in decreasing manner from the bottom that which Gene corresponds to a disease to occur and what is more likely there is a one to many relations. 

\begin{figure}
\scriptsize{
    \begin{lstlisting}[frame=bt,numbers=none,language=SPARQL,linewidth=\linewidth,morekeywords={FILTER,AS,GROUP,BY,VALUES,PREFIX}]
    SELECT DISTINCT str(?dName) as ?dName str(?gSymbol) as ?geneSymbol 
    count(?article) as ?publications
    WHERE {
    	?gda rdf:type ?type;
            sio:SIO_000628 ?gene,?disease;
     		sio:SIO_000772 ?article .
        ?disease a ncit:C7057;
            	dcterms:title ?dName .
    	?gene a ncit:C16612;
                sio:SIO_000205 ?symbolUri .
        ?symbolUri dcterms:title ?gSymbol .
    }
    ORDER BY DESC (?publications)
    LIMIT 20
\end{lstlisting}}
    \caption{Query for list of oncogenes responsible for any type of cancer cited in Cancer Index} 
    \label{fig:sparql_query_2}
\end{figure}

\subsubsection{Which genes are commonly found among diseases?}
\noindent Overlapping disease with a gene can answer questions when one analyzes the VDAs. Genetic Variation is what makes us unique. The modification in the DNA can help predict answer about one's appearance such as hair colour, skin colour or even the shape of our faces. 

\begin{figure}
\scriptsize{
    \begin{lstlisting}[frame=bt,numbers=none,language=SPARQL,linewidth=\linewidth,morekeywords={FILTER,AS,GROUP,BY,VALUES,PREFIX}]
    SELECT ?gda ?gene ?geneSymbol ?disease ?diseaseName
    	WHERE {
    		?gda sio:SIO_000628 ?gene,?disease .
    		?gene rdf:type ncit:C16612 ;
    			sio:SIO_000205 ?symbolUri .
        	?symbolUri dcterms:title ?geneSymbol .
    		?disease rdf:type ncit:C7057;
    			dcterms:title ?diseaseName
    	}
    	LIMIT 100
\end{lstlisting}}
    \caption{Query for list of oncogenes responsible for any type of cancer cited in Cancer Index} 
    \label{fig:sparql_query_3}
\end{figure}

\begin{figure*}
    \centering
    \includegraphics[width=0.8\textwidth]{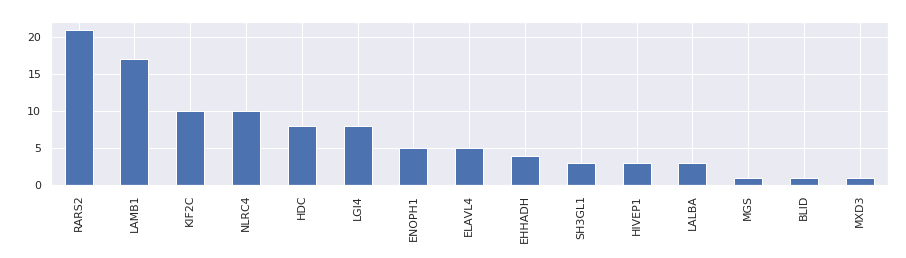}
    \caption{Common genes responsible for different diseases}
    \label{fig:commongene}
\end{figure*}

\subsection{Experiment setup}
\begin{figure*}
    \centering
    \includegraphics[width=0.8\textwidth]{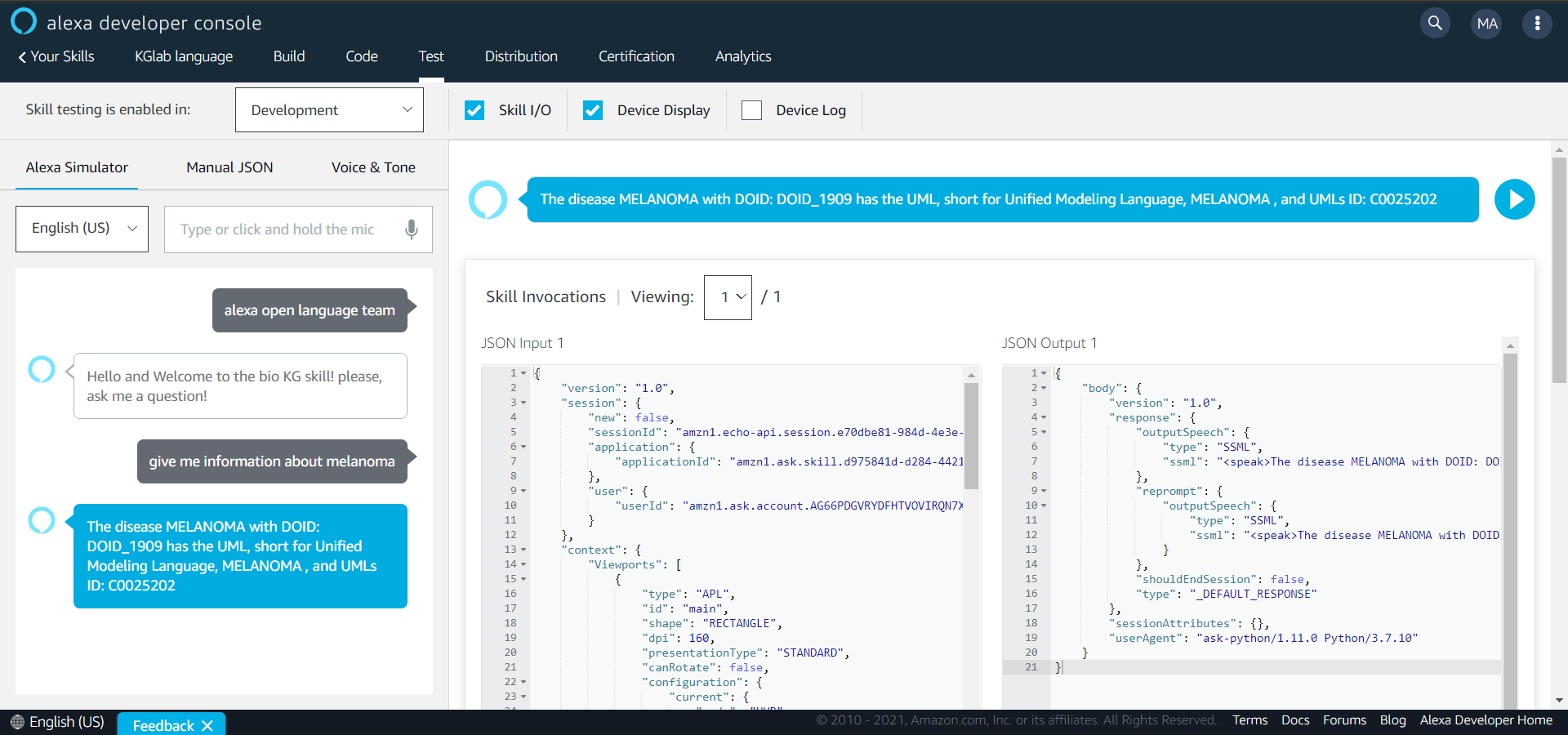}
    \caption{Testing of executing sample QA about specific disease over Alexa console}
    \label{fig:testing}
\end{figure*}

\noindent 
In order to test our skill after iteration of adding new features, and what kinds of testing are we doing, unit testing, performance testing, etc. The Amazon development console provides an interface for testing our skill, with both texting and speech capabilities, and for the rest of testing methodologies. \Cref{fig:testing} shows an example of executing sample QA about specific disease over Alexa console. 

A convenient choice was using Amazon development console, since the code is already there, which makes the deployment and synchronize easier and faster\footnote{Developing locally on an IDE of choice, though. However, we run into the problem that each time we want to test the code, need to fully deploy the skill, and wait for that the Amazon Lambda function and the Alexa skill to synchronize, even though all the files were locally. Only then we can use the testing interface, and leaving a big overhead.}, because the configuration files are already deployed. As Amazon development console code section was not the very developer friendly\footnote{Technically it was not well-responsive as it did not catch all errors.}, 
we took a combined approach: codes, functions and handlers were written locally in our IDE, followed by migrating them to Amazon development console only for deployment, enabling faster deployment. 

\subsection{Building example skills}
\noindent We start building the skills by define our intents based on initial basic intents provided by Alexa, such as `Cancel', `Stop', `Help'. We then need define our custom intents. Then for each intent, necessary utterances were written that suit that specific intent, and the slots that are needed for it too. Finally, we defined the handler code for respective utterances. The handlers are specific to each intent. For every intent, we create a specific handler that responds with the correct response. 
Besides, we use a special intent called `definition'. It is used when a user wants basic information for a specific disease, disease ontology ID~(DOID) and unified modeling language ID~(UMLID). 
Utterances for this intent where something like: \textit{`Give me information about {slot}.},  or \textit{`I want to know more about {slot}'}. The slot value was the name of the disease, where Alexa requires to know what slot diseases is it waiting for. This is done explicitly giving the slot values and embedding them into Alexa, we started by defining few disease for testing purposes, followed by importing as much diseases, yielding we have over 12,000 different slots across different diseases. 


The handler code is responsible for mapping entities in the KG with the correct SPARQL query, using the variables given as the slot values, such that the query itself is as dynamic as possible. Upon receiving the response as a JSON, we developed customized functions called `printers' in order to reformat the JSON into human-readable language, which will both shown and spoken out the user by Alexa. We did not write customized functions for speaking the output, as this is pre-handled by Alexa. In fact, we just need to call the function that speaks the output we give to it. 
Overall, building Alexa skills and interacting with KG invoves: calling the intent, mapping it to a handler for sending an API request, getting the response, and speaking out the response back to the user. 







\subsection{Results and observations from Alexa} 
\noindent For the Alexa part we used the above mention methodology to apply the interaction model between the user and the knowledge graph, and we will explore some of the results that we got and concluded from it. As a baseline, the idea of using Alexa as an interface for a knowledge graph works great If applied to straight forward or a simple graph, but as the complexity starts to grow bigger, and queries and relations becomes more convoluted. 

\begin{figure*}
    \centering
    \includegraphics[width=0.8\textwidth]{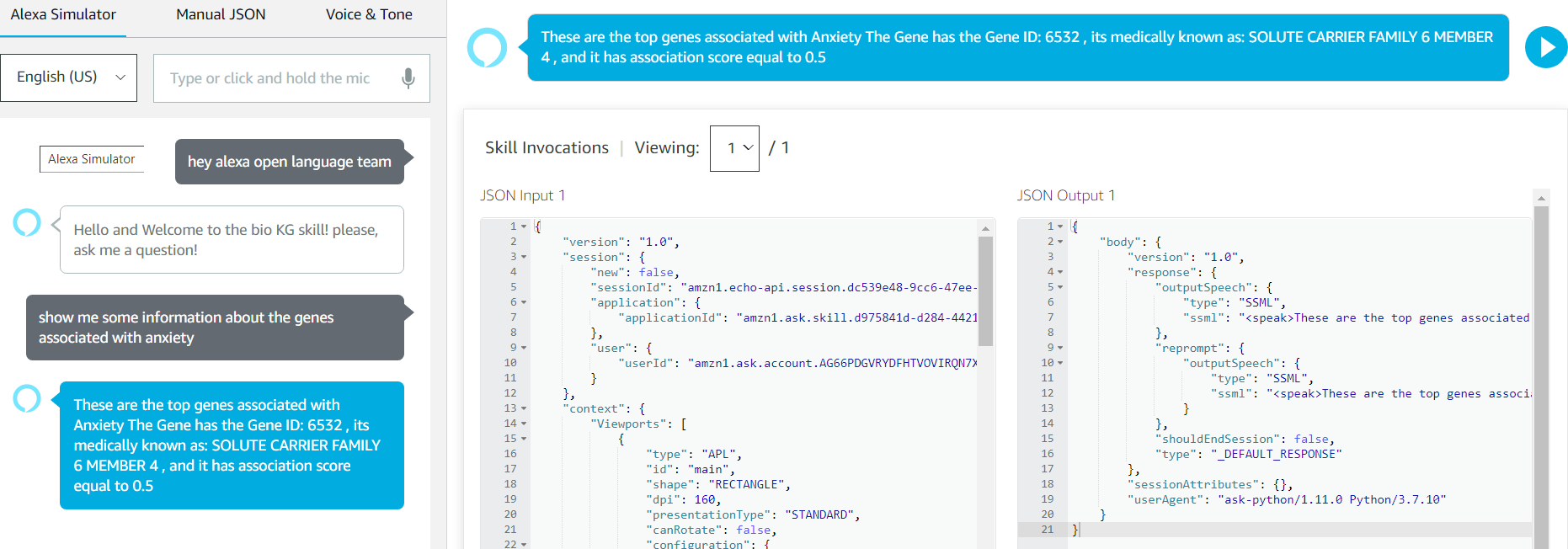}
    \caption{Testing of executing sample causation query about specific disease over Alexa console}
    \label{fig:causatoin}
\end{figure*}

To begin with we applied the simple disease definition intent, which in turn ran a simple information retrieval query, and then reads out the related disease identifiers out to the user. As shown in \cref{fig:testing}, our approach runs on any disease if we have it stored as a slot value, along with its corresponding disease ID. We then decided to increase the complexity, and the intent was a disease causation intent, this on tries to retrieve the top genes associated with a certain disease according to scoring metric in the knowledge graph. Running the query from the development console, we get the result represented to the user in a readable format. \Cref{fig:causatoin} presents an idea on how the intent and result from the backed KG. 

Moreover, we can apply we have to any other intent-query combination that exhibits a similar complexity. However, problems start to happen when increase query complexity to what we call multilayered query, and we will elaborate on that now. For both queries mentioned above, what we do is that we send one query as a one API call to the knowledge graph, it does not really matter how large the query itself is, or how large the response, as long as we do just one round trip to the knowledge graph and back. So when we try a multi layer query, what we mean is that we send the query to the back end, get the result, based on it we send another queries with the result we got back, and we can do that so on\footnote{For example, we can send a definition query about a disease, get the UMLID, use this result to get association score based some metric via another query, in this case it will be 2-layer query.}.  

\section{Conclusion and Outlook}\label{sec:con}
\noindent In this paper, we represent a proof-of-concept how to perform QA over KGs using Amazon Alexa's voice-enabled interface. We use the well-known \emph{DisgeNET} KG, which contain knowledge covering 1.13 million gene-disease associations between 21,671 genes and 30,170 diseases, disorders, and clinical or abnormal human phenotypes. Our study shows how Alex could be of help to find facts about certain biological entities from large-scale knowledge bases. We explored how we can we exploit a medical knowledge graph DisGeNET, and integrate it to a voice-based interface, in this case Alexa and create a skill by integrating it through SPARQL endpoint and placing the RDF queries in the custom handlers of the skill.

This study also report some technical issues. For example, the queries work perfectly from a local IDE, however they fail to execute properly in the Alexa developer console. The possible reasons might be cloud console's code formatting problem which the Alexa console fail to compile and execute. A second reason might be due to second or third level complexity and layers in the queries, the execution gets timed out. Further, in the current phase of the implementation, every time we send any request more than a 2-layered query, the skill crashes and it does not return anything, thereby we are stuck at this layer threshold of only 2, which severely limits how deep and complex we can make the intents. A possible solution is to make the queries as complex as possible and put the responsibility of graph crawling on the query itself, but this is horribly inefficient, very hard to trace and scale. It is important to note that we did not run into this problem when we ran the code locally. We can create a very arbitrarily deep layered query, which is simple to trace and construct, run it on the knowledge graph, and we get the desired result. The problem only occurs when we migrate the code into the Alexa development console, and try to test the interface from it, only then that skill crashes.
   
In the future, we intend to overcome these issues by integrating over multiple knowledge graphs and increase the performance~\cite{10.1145/3366423.3380289} to create a more general knowledge base which will help researchers in the medical field. Further, we look forward to implement few more queries by collaborating with natural language understanding researchers. 

\section*{Acknowledgement}
This paper is based on the Knowledge Graph Lab course\footnote{https://dbis.rwth-aachen.de/dbis/index.php/} offered at Computer Science 5 - Information Systems and Databases, RWTH Aachen University, Germany and a joint collaboration with Osthus GmbH\footnote{https://www.osthus.com/}, Aachen, Germany. 

\bibliographystyle{IEEEtran}
\bibliography{references.bib}

\end{document}